\documentclass[10pt,twocolumn,letterpaper]{article}

\usepackage{iccv}
\usepackage{times}
\usepackage{epsfig}
\usepackage{graphicx}
\usepackage{animate}
\usepackage{amsmath}
\usepackage{amssymb}
\usepackage{pifont}
\usepackage{lipsum} 
\usepackage{color, colortbl, xcolor}
\usepackage{multirow}
\definecolor{Gray}{gray}{0.9}
\usepackage{subcaption}
\newcommand{\xmark}{\scriptsize\ding{53}}%

\usepackage[group-separator={,}]{siunitx} 
\usepackage[shortlabels]{enumitem} 
\usepackage{booktabs} 
\usepackage{tikz} 
\usepackage{pgfplots} 
\pgfplotsset{compat=newest}
\usepackage[accsupp]{axessibility}  

\usepackage[pagebackref=true,breaklinks=true,letterpaper=true,colorlinks,bookmarks=false]{hyperref}

\iccvfinalcopy 

\def\iccvPaperID{1106} 

\ificcvfinal\fi

\begin{document}
\newcommand{\mypartop}[1]{\vspace{0mm}\noindent\textbf{#1}}
\newcommand{\mypar}[1]{\vspace{0.5em}\noindent\textbf{#1}}

\newcommand{\coco}{MS~COCO}
\newcommand{\cococite}{\cite{chen2015microsoft,mscoco}}
\newcommand{\flickr}{Flickr30k Entities}
\newcommand{\flickrcite}{\cite{young14tacl,plummer17ijcv}}

\newcommand{\taskname}{Panoptic Narrative Grounding}
\newcommand{\com}{CoM}

\newif\ifdraft
\drafttrue
\ifdraft
  \newcommand{\Jordi}[1]{{\color{red}[Jordi] #1}\xspace}
  \newcommand{\Cristina}[1]{{\color{violet}[Cristina] #1}\xspace}
  \newcommand{\Isabela}[1]{{\color{green}[Isabela] #1}\xspace}
  \newcommand{\att}[1]{{\color{magenta}#1}}
\else
  \newcommand{\Jordi}[1]{}
  \newcommand{\Cristina}[1]{}
  \newcommand{\Isabela}[1]{}
  \newcommand{\att}[1]{#1}
\fi

\title{\taskname{}}

\author{Cristina Gonz\'{a}lez$^1$$\qquad$Nicolás Ayobi$^1$$\qquad$Isabela Hern\'{a}ndez$^1$$\qquad$Jos\'{e} Hern\'{a}ndez$^1$\\Jordi Pont-Tuset$^2$$\qquad$Pablo Arbel\'{a}ez$^1$\\[2mm]
    $^1$Center for Research and Formation in Artificial Intelligence, Universidad de los Andes, Colombia\\[2mm]$^2$Google Research, Switzerland
}

\maketitle
\ificcvfinal\thispagestyle{empty}\fi

\begin{abstract}
This paper proposes \emph{\taskname{}}, a spatially fine and general formulation of the natural language visual grounding problem.
We establish an experimental framework for the study of this new task, including new ground truth and metrics,
and we propose a strong baseline method to serve as stepping stone for future work.
We exploit the intrinsic semantic richness in an image by including panoptic categories, and we approach visual grounding at a fine-grained level by using segmentations.
In terms of ground truth, we propose an algorithm to automatically transfer Localized Narratives annotations to specific regions in the panoptic segmentations of the \coco{} dataset.
To guarantee the quality of our annotations, we take advantage of the semantic structure contained in WordNet to exclusively incorporate noun phrases that are grounded to a meaningfully related panoptic segmentation region.
The proposed baseline achieves a performance of \num{55.4} absolute Average Recall points.
This result is a suitable foundation to push the envelope further in the development of methods for \taskname{}.
\end{abstract}

\section{Introduction}\label{sec:introduction}
Vision and language skills play a key role in humans' understanding of the world and they are rarely used independently.
Their interaction is crucial in achieving high-level tasks such as describing objects, narrating a visual scene,
or answering questions based on visual cues.
Inspired by these capabilities of human intelligence, researchers have formulated tasks at the intersection of computer vision and natural language processing, such as image captioning~\cite{chen2015microsoft,stanfordvp,mscoco,conceptual_captions}, referring expression comprehension and segmentation~\cite{kazemzadeh2014referitgame,krishna2017visual,refexp,mao2016generation}, visual question answering~\cite{antol2015vqa}, among many others.

\begin{figure}[t]
\begin{center}
  \includegraphics[width=\linewidth]{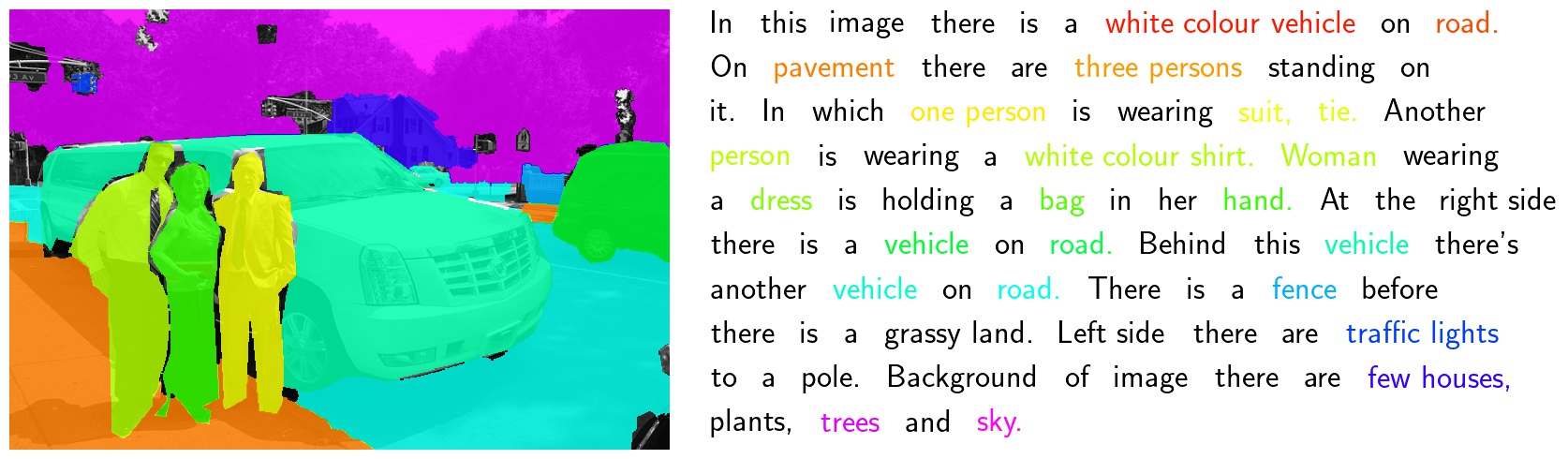}
\end{center}
  \caption{\textbf{\taskname{}}. Given an input image (left) with an associated caption (right), our goal is to produce a panoptic segmentation that grounds its visual objects densely (left). }
 \label{fig:task}
\end{figure}

Current experimental frameworks approach vision and language tasks at different granularity levels.
Image captioning is among the coarsest, aiming at pairing an image with a textual description of its content.
With increasing granularity, there are frameworks that intend to assign specific regions in the image to short descriptions,
such as referring expression comprehension and segmentation, and region descriptions in the
Visual Genome~\cite{krishna2017visual} dataset.
The finest approaches tackle visual grounding at word level, with bounding boxes in the image linked to noun phrases in
the caption, as in the \flickr{}~\flickrcite{} dataset.

Given the general dichotomy in computer vision tasks between \emph{things} (countable objects) and \emph{stuff} (amorphous regions of similar texture), these datasets mainly focus on things categories.
However, several works~\cite{forsyth1996finding,adelson2001seeing,heitz2008learning,kirillov2019panoptic} have highlighted the importance of jointly considering both things and stuff classes towards real-world applications.
Kirillov~\etal\cite{kirillov2019panoptic} defined the panoptic segmentation task as the unified
formulation of semantic segmentation, which recognizes \emph{stuff}, and instance segmentation, which
detects and segments \emph{things}.

Recently, Pont-Tuset~\etal\cite{pont2020connecting} proposed Localized Narratives, a multimodal dataset for visual grounding based on the natural human task of describing images while simultaneously pointing with the mouse at the regions being described.
Their grounding annotations have the densest sampling of current datasets, in that they visually ground every word in the caption with mouse trace segments.
Moreover, this dataset naturally contains panoptic descriptions of the content of each image, including \emph{stuff} regions.

Although some of the experimental frameworks above contain sufficiently dense annotations over language, visual annotations are still very sparse and rough.
This paper proposes \emph{\taskname{}} (Figure~\ref{fig:task}), a spatially finer and more general task formulation of the natural language visual grounding problem in which (i) we propose a spatially detailed visual grounding that uses segmentations instead of bounding boxes, and (ii) we include all panoptic categories to fully exploit the intrinsic semantic richness in a visual scene.
We establish an experimental framework for the study of this task, including new ground-truth annotations and metrics, and we propose a strong baseline to serve as stepping stone for future work.

Considering that collecting pixel-wise annotations has a significant cost, we design a method to transfer the Localized Narrative annotations to regions into the panoptic segmentation annotations provided by the \coco{}~\cococite{} dataset.
We select a specific region for each noun phrase considering the trace segment associated to it.
Since humans use different styles to pointing objects (circling, scribbling, and underlining)
we evaluate the quality of the annotation by determining if noun phrases are grounded to a meaningfully related panoptic segmentation region.
To do so, we leverage the underlying semantic information in the WordNet ontology~\cite{miller1998wordnet}.
To handle trace segments being not fully synchronized with the object description,
we consider all the regions in the image ranked by relative distance to the visual grounding and assign the closest region with a strong meaning relationship.
Finally, for plural noun phrases, we select all regions that (i) are within the tightest bounding box around the mouse trace and
(ii) are from the same category as the main selected region.
In terms of evaluation, in contrast to the traditional metric use in phrase-grounding task,
we introduce a stricter metric that calculates recall at different Intersection over Union (IoU) thresholds between the target and the predicted mask.
Thus, measuring both recognition accuracy and segmentation quality.

We propose a strong baseline that builds upon a state-of-the-art method, Cross-Modality Relevance~\cite{zheng2020cross} (CMR), developed for reasoning tasks on language and vision.
Specifically, this method is designed to perform two classification tasks: Visual Questions Answering~\cite{antol2015vqa} and Natural Language for Visual Reasoning~\cite{suhr2017corpus,suhr2018corpus}.
We generalize the model to perform \taskname{}.
Thus, our baseline is the first natural language visual grounding method able to align multiple noun phrases with panoptic region segmentations within the image.

Our main contributions can be summarized as follows:
\begin{enumerate}[(1),nosep,leftmargin=*,widest*=8]
    \item We propose \taskname{}, a new formulation of the natural language visual grounding problem which, by using panoptic segmentation regions as visual grounding, is spatially denser and more general in semantic terms.
    \item We establish an experimental framework for the study of this problem, with annotations coming from the transfer of Localized Narrative annotations to panoptic segmentations in the~\coco{} dataset.
    \item We introduce the first visual grounding method that matches segmentation regions to specific noun phrases in the caption, which serves as a strong baseline for the task of \taskname{}.
\end{enumerate}

To ensure the reproducibility of our results and to promote further research on \taskname{}, we make all the resources of this paper publicly available in our project web page\footnote{\url{https://github.com/BCV-Uniandes/PNG}}: our benchmark dataset annotations for the train and validation splits in \coco{}, an implementation of the evaluation metrics, and the pretrained models and source code for our baseline.

\section{Related Work}\label{sec:related-work}
\subsection{Vision and Language Datasets}\label{subsec:datasets}
The first datasets at the intersection of vision and language match images to descriptions without any form of visual grounding~\cite{chen2015microsoft,stanfordvp,mscoco,conceptual_captions}.
\coco{} Captions~\cococite{} comprises \num{123287} images for training, validation, and testing;
with five human-annotated captions for each one.
Several works, however, have suggested that models developed for this task do not base their decisions on the visual information contained in the image, but rather on easier-to-learn language priors~\cite{selvaraju2019taking}.

Referring Expressions Comprehension (REC) and Referring Expression Segmentation (RES) are tasks intended to detect and segment, respectively, a target object instance described by a natural language expression.
There exist three datasets for these tasks, built on top of the \coco{} dataset: RefCOCO~\cite{kazemzadeh2014referitgame}, RefCOCO+~\cite{kazemzadeh2014referitgame}, and RefCOCOg~\cite{mao2016generation}.
RefCOCO+'s descriptions, in contrast to those on RefCOCO, focus on appearance attributes rather than locations of the referent object.
RefCOCOg was collected in a non-interactive setup and contains longer descriptions including both appearance and location attributes of the target instances.
These datasets, while having pixel-level
annotations, do not ground language at the word-level,
and their annotations' semantics boil down mostly to salient objects in images.

Visual Genome~\cite{krishna2017visual} is a general-purpose dataset meant to connect vision and language while not being restricted to a specific task.
It contains \num{108077} images with \num{50} region descriptions and \num{35} objects per image on average.
Each short description is localized within the image with a bounding box, making it possible to study visual grounding at the phrase level.
There is no an explicit relationship between the objects localized within an image and the visual entities included in the textual descriptions. 
Having multiple regions descriptions allows to have a more complete understanding of a scene, covering \emph{things} and \textit{stuff} categories.

\flickr{}~\cite{plummer17ijcv} extends the original Flickr30k dataset~\cite{young14tacl} with manually annotated bounding boxes grounding each noun phrase in the caption.
Despite its granularity over language, this dataset does not provide pixel-wise annotations and focus predominantly on \emph{things}.

Localized Narratives~\cite{pont2020connecting} is a large-scale dataset that provides annotations for the whole \coco{}, Flickr30k~\cite{young2014image}, and ADE20K~\cite{zhou2017scene} datasets, and \num{671}K images from Open Images~\cite{kuznetsova2020open}.
Its annotations consist of synchronized voice recordings, text transcriptions, and mouse traces made by human annotators.
They describe the image's contents and the mouse traces visually ground each word in the narrative, which makes them the densest annotations over language available to date.
Grounding goes beyond nouns and includes visual relationship indicators, verbs, \etc.
Annotators were asked to make descriptions that comprise as much content of the image as possible, which generally entails a high semantic coverage of the scene.
The visual grounding using mouse trace segments, however, is spatially very coarse.

Table~\ref{tab:datasets} summarizes the characteristics of existing datasets at the intersection of vision and language and compares them in terms of: (i) language granularity, (ii) visual granularity, and (iii) semantic generality.
Our benchmark dataset is designed to fill the gaps between all previous datasets:
(i) It maintains the finest granularity over language by grounding specific words in the narrative, as the latest developed natural language grounding datasets.
(ii) It provides fine-grained visual grounding annotations using segmentations, as in referring expression segmentation datasets.
This allows studying the problem with a stricter experimental framework in which the location on the objects is not approximate as in the object detection task.
(iii) Our annotations include \emph{panoptic} categories~\cite{kirillov2019panoptic} instead of just focusing on \emph{things} categories,
which gives way to a global analysis of the scene which is relevant for reasoning about the world around us.

\subsection{Natural Language Visual Grounding Methods}\label{subsec:methods}
The methods developed for the coarsest natural language visual grounding tasks on language, specifically for the REC and RES tasks, can be grouped into two general approaches: top-down and bottom-up.
The former paradigm~\cite{chen2019referring,liu2019learning,yu2018mattnet} starts from region proposals extracted by a general method of object detection or segmentation.
From this set of objects, the method selects the one that is described to a greater extent by the natural language expression.
The strengths of this paradigm are that they can take advantage of what general object detection and segmentation methods have learned from the existing large-scale datasets for these general tasks,
rather than directly learning how to detect objects within an image.
However, since these methods do not update the region proposals,
the upper bound of these methods is significantly impacted by the performance of the base method.
In contrast, the latter approaches produce the referred object segmentation by grouping pixels~\cite{chen2019see,hu2016segmentation,hu2020bi,huang2020referring,hui2020linguistic,li2018referring,li2017tracking,liu2017recurrent,luo2020cascade,luo2020multi,margffoy2018dynamic,qiu2019referring,rong2019unambiguous,shi2018key,ye2020dual,ye2019cross}.
These methods use single networks, which leverage high and low-level features to refine segmentation masks along levels of their architectures.
However, complying with a larger search space than its top-down counterpart, leads to higher rates of false positive segmentations. 

Some existing methods approach phrase grounding through a combination of base networks for visual and linguistic information.
To this end, Convolutional Neural Networks (CNN) and Recurrent Neural Networks (RNN) are used~\cite{hu2016segmentation,margffoy2018dynamic}, and their features leveraged to achieve word-to-image interactions. 
Other works apply Transformer attention~\cite{vaswani2017attention} to generate contextualized representations of both modalities. 
The latter architectures are also used in cross-modality analyses to aggregate and align visual cues and linguistic meanings~\cite{visualbert,li2020does,vilbert,vlbert,zheng2020cross}.

\begin{table}[!t]
\resizebox{\linewidth}{!}{%
\begin{tabular}{c c c c}
\toprule
Dataset & \begin{tabular}[c]{@{}c@{}}Language\\ Granularity\end{tabular} & \begin{tabular}[c]{@{}c@{}}Visual\\ Granularity\end{tabular} & \begin{tabular}[c]{@{}c@{}}Semantic\\ Generality\end{tabular} \\ \midrule
\coco{} Captions~\cococite{} & Caption & \xmark & \xmark \\
\rowcolor{Gray} Conceptual Captions~\cite{conceptual_captions} & Caption & \xmark & \xmark \\
Stanford Visual Par.~\cite{stanfordvp} & Caption & \xmark & \xmark \\
\rowcolor{Gray} ReferIt~\cite{kazemzadeh2014referitgame,mao2016generation} & Short phrase & Segmentation & Things\\
Google Refexp~\cite{refexp} & Short phrase & Segmentation & Things\\
\rowcolor{Gray} Visual Genome~\cite{krishna2017visual} & Short phrase & Bounding box & Things $+$ Stuff \\
\flickr{}~\flickrcite{} & Noun phrase & Bounding box & Mainly Things \\ 
\rowcolor{Gray} Localized Narratives~\cite{pont2020connecting} & Each word & Traces & Things $+$ Stuff \\
\midrule
\textbf{\taskname{} (Ours)} & \textbf{Noun phrase} & \textbf{Segmentation} & \textbf{Things $+$ Stuff} \\
    \bottomrule
\end{tabular}%
}
\vspace{1mm}
\caption{\taskname{} compared with major captioning and natural language grounding datasets.}
\label{tab:datasets}
\end{table}

\section{\taskname{} Benchmark}\label{sec:dataset}

\begin{figure*}[!ht]
    \centering
    \includegraphics[width=0.83\textwidth]{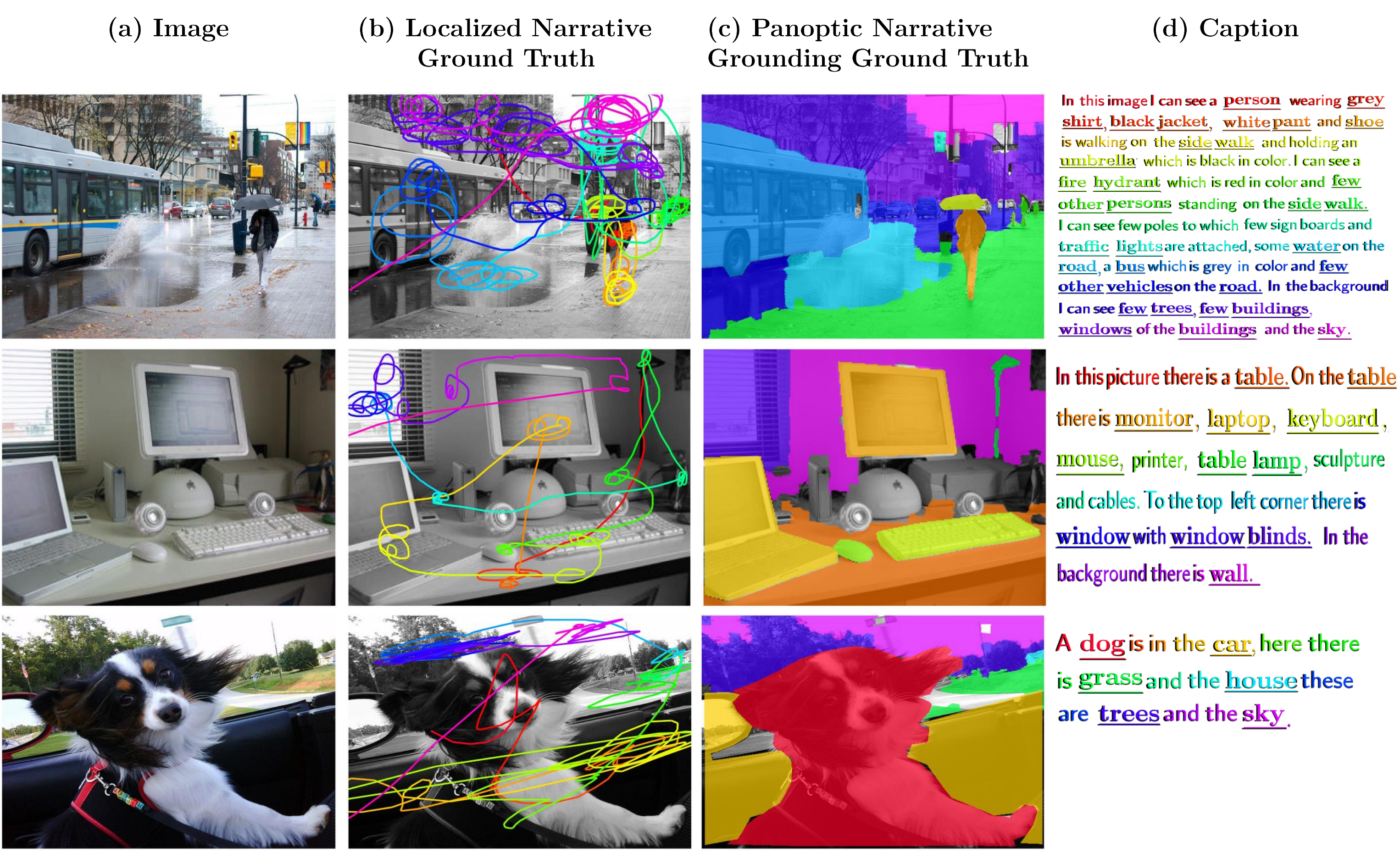}
    \caption{\textbf{Ground-Truth Annotations Results.} Examples of different \taskname{} ground truth resulting from the proposed annotation transfer algorithm (c). We show the input image (a) and Localized Narrative traces (b) and caption with the matched panoptic segmentation regions (d). Color gradient in the trace, panoptic segmentation and caption indicates time over the language. The segmentation regions are visualized with the color of their corresponding noun phrases, according to the last associated spoken word. More qualitative results can be found in the Supplementary Material.
    }
    \label{fig:general_matches}
\end{figure*}

This section describes our proposed benchmark.
We describe how we generate the ground-truth annotations (Sec.~\ref{subsec:annotation-transfer}) and report some statistics (Sec.~\ref{subsec:dataset-statistics}).
Finally, we explain the proposed metrics to evaluate results on our benchmark (Sec.~\ref{subsec:metrics}).

\subsection{Ground-Truth Annotations}\label{subsec:annotation-transfer}
We transfer the Localized Narrative~\cite{pont2020connecting} annotations to~\coco{} panoptic segmentations by synchronizing timed captions and trace points.
We consider a set of utterances $U=\{u_1,\dots,u_n\}$ in the caption and a set of trace points $T=\{t_1,\dots,t_m\}$ for each Localized Narrative, where $u_i$ and $t_j$ are the timestamped verbal and spatial units that, respectively, compose a caption and a mouse trace.

Synchronization occurs by selecting the trace segments that are drawn over an image between the start and end times of an utterance, or a subset of contiguous utterances, containing a noun phrase $(u_{in})$.
Each $u_{in}$ is hence associated with a subset of trace points $T'=\{t_a,\dots,t_b\}$ (where $t_a$ and $t_b$ are within the timestamp of $u_i$). $T'$ provides a spatial reference for each utterance through their point coordinates.
We average these coordinates to obtain a single point $p$ in the image plane, which we refer to as the Center of Mass (\com{}).
In the Supplementary Material, we exemplify how this strategy of summarizing the annotator's grounding is beneficial for irregular and circling patterns in the Localized Narrative annotations. 

To identify noun phrases, we perform a chunking parsing using the Natural Language Toolkit (NLTK) for Python~\cite{nltk}, and select sequences of consecutive nouns that may or may not have a previous adjective or a cardinal digit. These noun phrases are considered a single grounding unit.

Let $S=\{s_1,\dots,s_k\}$ be the regions from the \coco{} panoptic segmentation of the same image, with corresponding \emph{thing} or \emph{stuff} category labels $C=\{c_1,\dots,c_k\}$.
We select the region $s_i\in S$ which contains the location of the \com{} $p$ of the mouse trace and establishes it as a candidate for the ground truth of the noun phrase $u_{in}$.
This matching, however, is not guaranteed to be correct, as Localized Narrative mouse traces might not fall exactly on the grounded object.
We, therefore, filter these matches by comparing the noun phrase in the caption ($u_{in}$) to the \coco{} object category associated with $s_i$ ($c_i$), to which should \textit{agree}.
Intuitively, if the annotator said $u_{in}=$``red vehicle'', we want to consider the matching correct if \eg $c_i=$``vehicle'' or $c_i=$``car'';
but we want to discard it if $c_i=$``tree'', since it might mean that the annotator was pointing to a tree next to a vehicle when they pronounced the word ``vehicle''.
Furthermore, if the caption refers to the same object in the visual scene multiple times, our benchmark will associate it with all the related noun phrases.
Figure~\ref{fig:general_matches} shows some examples of our visual grounding annotations.

To evaluate whether the caption utterance(s) $u_{in}$, or its composing noun(s), correspond to the matched panoptic category $c_i$,
we consider the following criteria:
(1) \textit{exact matches} ($u_{in}$ is strictly equal to $c_i$),
(2) \textit{synonyms} ($u_{in}$ is a synonym of $c_i$),
(3) \textit{hierarchical relationship} ($u_{in}$ is either a hypernym or hyponym of $c_i$), and
(4) \textit{meronyms} ($u_{in}$ is a meronym of $c_i$).
We evaluate (1) by simple string comparison, and (2), (3) and (4) accessing WordNet~\cite{miller1998wordnet} using NLTK~\cite{nltk}.
Additionally, (5) we \textit{manually} relate specific words to certain \coco{} categories that are omitted by the WordNet ontology. Examples of these words include clothing pieces, body parts and female figures, as members of the \coco{} ``person'' category. 
We consider (1) as the highest rank in matching, given the similarity between words in this level, followed by (2), (3), (4) and (5).
Tag clouds are reported in the Supplementary Material as examples of criteria (2) - (4). 

We find two special cases that require additional measures when transferring annotations.
First, there is a possibility of time shifts~\cite{pont2020connecting}: an annotator moving their mouse slightly before or after describing objects.
To address this issue, we consider neighboring segmentation regions of the selected one via $p$ (the \com{} of the mouse traces), as potential matches to each noun phrase $u_{in}$. 
If no matching occurs with the center region, we proceed to select the closest candidate that matches $u_{in}$, with any of the semantic relationships defined above, as the grounding annotation for the noun phrase. 
We consider the minimum distance between segmentation regions as a measure of their closeness (following a \textit{single-linkage} concept). 
An example of vicinity region analysis can be found in the Supplementary Material.

\begin{figure}[t]
    \centering
    \includegraphics[width=0.9\linewidth]{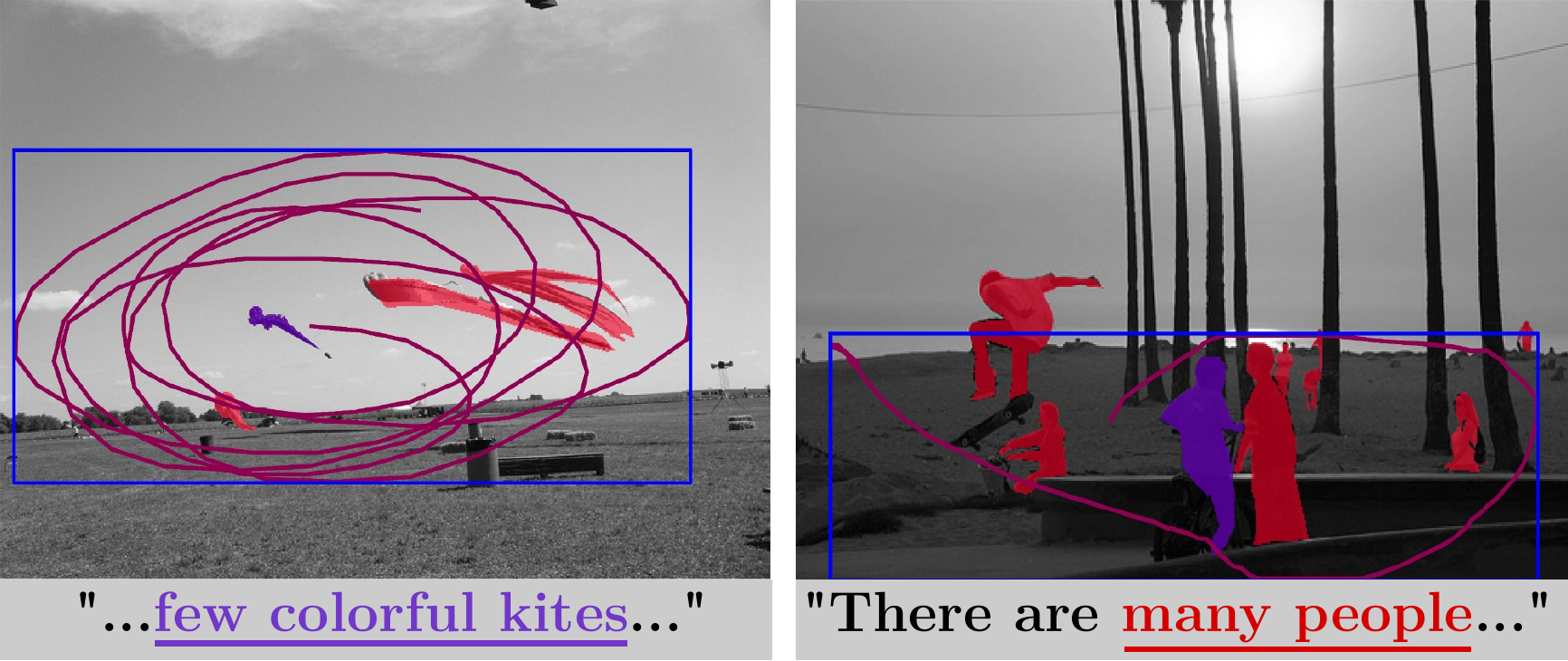}
    \caption{\textbf{Examples of grounded plural noun phrases.} We match multiple regions (red) beyond the seed region (purple).
    Increasing the spatial scope of our annotations agrees with the multi-instance nature of plural noun phrases and better captures the annotators’ intention of referring to several instances. Best viewed in color.}
    \label{fig:plural_matches}
\end{figure}

Second, we consider the implications of plural noun phrases within narratives.
These phrases are inherently related to various instances of an object, which are all globally pointed at during the utterance(s).
This motivates the mapping of a plural noun phrase to $S'\subseteq S$, which contains several regions with a common category, as opposed to a single segmentation $s_i$.
The common category is defined using $p$ and the vicinity region analysis, from which we select a \emph{seed} region.
We then augment this set with all the regions of the same category that are contained in the tightest bounding box around the mouse trace $T'$.
All regions in $S'$ are considered the grounding of the plural noun phrase $u_{in}$.
Figure~\ref{fig:plural_matches} shows examples of plural groundings.

\begin{figure*}[!t]
    \centering
    \includegraphics[width=0.99\textwidth]{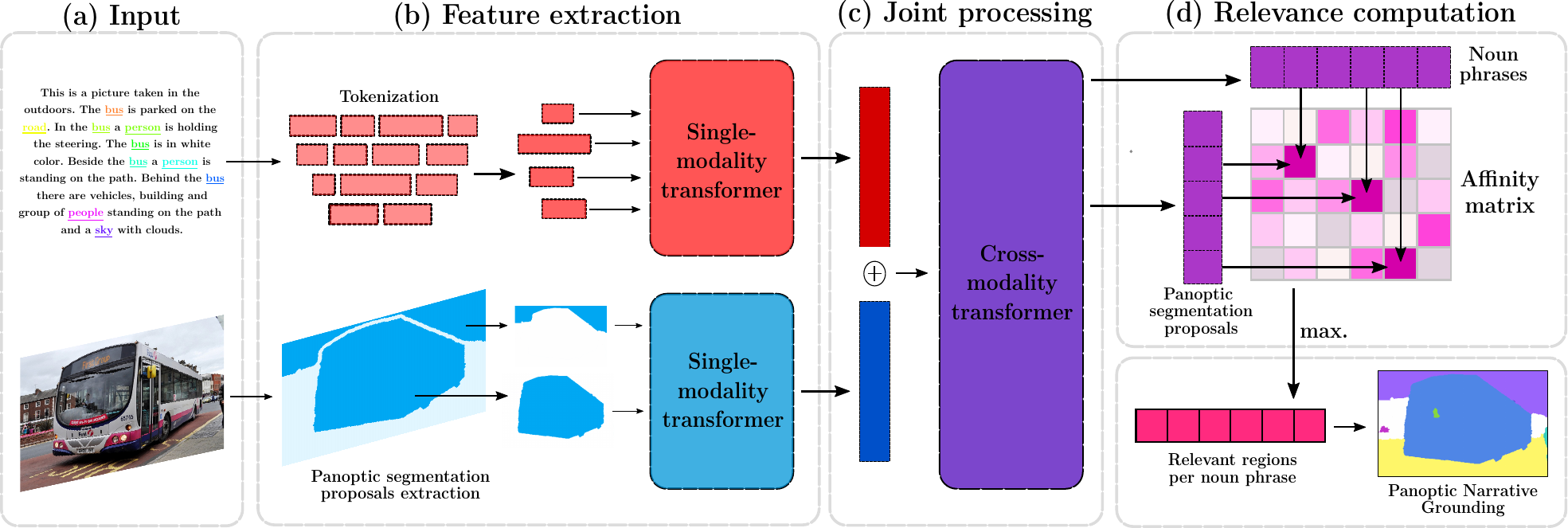}
    \caption{\textbf{\taskname{} Baseline.} Our model takes an image and a corresponding caption as input (a), followed by a top-down approach by extracting panoptic segmentation proposals with a backbone segmenter.
    We then use single-modality transformers to independently process the segmentation proposals and the tokens from the caption (b) and jointly process the information with a cross-modality transformer (c).
    Finally, we calculate an affinity matrix between visual and textual information and select the more relevant panoptic segmentation proposal for each noun phrase (d).
    }
    \label{fig:method}
\end{figure*}

\subsection{Dataset Statistics}\label{subsec:dataset-statistics}
Our complete annotation transfer methodology matches an average of \num{5.1} noun phrases per narrative and \num{726445} noun phrases in the whole Localized Narratives annotations for \coco{} dataset. Localized Narratives contain an average of  \num{11.3} noun phrases per narrative.
This accounts for \num{45.1}\% of noun phrases coverage, which we ground through panoptic segmentation annotations.
In the visual domain, this translates to a total of \num{741697} segments matched, from which \num{659298} are unique. Thus, our benchmark covers \num{47.5}\% of all the segments annotated in \coco{} panoptic with at least one noun phrase match.
 In turn, this coverage accounts for \num{58.5}\% of all pixels in \num{123287} \coco{} images and \num{65.4}\% relative to all annotated pixels in \coco{} panoptic annotations.

The proposed \taskname{} dataset ground both \textit{things} and \textit{stuff} categories.
Specifically, \num{57.0}\% of the noun phrases grounded are \textit{things} and \num{43.0}\% are \textit{stuff}.
\num{64.7}\% of matches between grounded noun phrases and \coco{} panoptic categories are exact matches between the category name and a noun in the phrase. Thus, using synonyms, manual matching, and hierarchical and menoronyms relationships significantly extend the grounding scope (\num{35.3}\%) and allow us to better capture the diversity in natural language expressions in Localized Narratives.
Lastly, \num{29.3}\% of the matched noun phrases correspond to plurals and \num{44.6}\% of matches are due to the vicinity region analysis. Consequently, each of the proposed matching steps plays an important role in improving the visual grounding annotations. A detailed report of these statistics can be found in the Supplementary Material.
Moreover, we carry out a final manual curation to verify the integrity of our annotations in the 1\% of the annotations.

\subsection{Metrics}\label{subsec:metrics}
In contrast to the recall metric, traditionally used for phrase grounding, we propose to calculate the Average Recall.
This metric evaluates the performance of the method in the \taskname{} task by considering different Intersection over Union (IoU) thresholds between the panoptic segmentation proposal and the ground-truth for each noun phrase.
Hence, the quality of the segmentations affects the method's performance since the IoU metric determines whether a detection is considered a true positive or not.
We obtain a curve that at very low IoU values has recall close to one and at higher IoU values recall drops.
The final metric, named Average Recall, is the area under the curve described above.
For plural noun phrases, we do not explicitly match the ground-truth instances and the prediction proposals to calculate the Average Recall.
Instead, we aggregate all instances of the ground-truth annotation into a single segmentation and compute the IoU with respect to the segmentation composed by all the prediction proposals.
With this strategy, we avoid errors or variations in the matching process between annotations and predictions while assessing the overall segmentation quality.

\section{Baseline for \taskname{}}\label{sec:baseline}
We build upon the Cross-Modality Relevance (CMR)~\cite{zheng2020cross} model developed for reasoning
tasks on language and vision.
This method has competitive results in two classification tasks at the intersection of vision
and language: Visual Question Answering~\cite{antol2015vqa} and Natural Language
for Visual Reasoning~\cite{suhr2017corpus,suhr2018corpus}.
CMR model introduces entity relevance representation that explicitly expressed the relevance of textual entities with respect to visual entities.
The model uses this affinity matrix as an intermediate representation for the final tasks.
Their results suggest using the word alignment with regions in the image results in an increase in performance for the final task.
We adapt the model and optimize the architecture for our task.

Figure~\ref{fig:method} depicts an overview of our baseline method.
Given an image and its caption (a),
we extract features from each of the panoptic segmentation region proposals and process each proposal and word with single modality transformers. (b).
We then concatenate the outputs from the visual modality transformer and the textual modality transformer and use it as input to the cross-modality transformer that updates the representations of each entity considering not only those of the same modality but also those of the other modality (c).
Afterwards, we calculate the affinity matrix from the matrix multiplication between the representations of each word and region proposal. Finally, we make an average pooling over the language dimension of all the words included in each noun phrase.
This affinity matrix explicitly indicates which is the most relevant panoptic segmentation region for each noun phrase in the caption (d).

\begin{figure*}[!ht]
   \centering
   \medskip
   \begin{subfigure}[t]{.33\linewidth}
  \centering
  \resizebox{0.95\linewidth}{!}{%
  \begin{tikzpicture}[/pgfplots/width=1.45\linewidth, /pgfplots/height=1.45\linewidth]
    \begin{axis}[
                 ymin=0,ymax=1,xmin=0,xmax=1,
        		 xlabel=IoU,
        		 ylabel=Recall@IoU,
         		 xlabel shift={-2pt},
        		 ylabel shift={-3pt},
		         font=\small,
		         axis equal image=true,
		         enlargelimits=false,
		         clip=true,
        	     grid style=solid, grid=both,
                 major grid style={white!85!black},
        		 minor grid style={white!95!black},
		 		 xtick={0,0.1,...,1.1},
                 xticklabels={0,.1,.2,.3,.4,.5,.6,.7,.8,.9,1},
        		 ytick={0,0.1,...,1.1},
                 yticklabels={0,.1,.2,.3,.4,.5,.6,.7,.8,.9,1},
         		 minor xtick={0,0.02,...,1},
		         minor ytick={0,0.02,...,1},
        		 legend style={at={(0.05,0.05)},
                 		       anchor=south west},
                 legend cell align={left}]
    \addplot+[red,dashed,mark=none,ultra thick] table[x=IoU,y=Oracle]{figures/figure1_final.txt};
    \addlegendentry{Oracle}
    \addplot+[red,solid,mark=none,ultra thick] table[x=IoU,y=Ours]{figures/figure1_final.txt};
    \addlegendentry{Ours}
    \end{axis}
\end{tikzpicture}}
  \subcaption{Overall performance}
  \label{fig:overall}
\end{subfigure}
\begin{subfigure}[t]{.33\linewidth}
  \centering
    \resizebox{0.95\linewidth}{!}{%
  \begin{tikzpicture}[/pgfplots/width=1.45\linewidth, /pgfplots/height=1.45\linewidth]
    \begin{axis}[
                 ymin=0,ymax=1,xmin=0,xmax=1,
        		 xlabel=IoU,
        		 ylabel=Recall@IoU,
         		 xlabel shift={-2pt},
        		 ylabel shift={-3pt},
		         font=\small,
		         axis equal image=true,
		         enlargelimits=false,
		         clip=true,
        	     grid style=solid, grid=both,
                 major grid style={white!85!black},
        		 minor grid style={white!95!black},
		 		 xtick={0,0.1,...,1.1},
                 xticklabels={0,.1,.2,.3,.4,.5,.6,.7,.8,.9,1},
        		 ytick={0,0.1,...,1.1},
                 yticklabels={0,.1,.2,.3,.4,.5,.6,.7,.8,.9,1},
         		 minor xtick={0,0.02,...,1},
		         minor ytick={0,0.02,...,1},
        		 legend style={at={(0.05,0.05)},
                 		       anchor=south west},
                 legend cell align={left}]
    \addplot+[orange,dashed,mark=none,ultra thick] table[x=IoU,y=Oracle_things]{figures/figure2_final.txt};
    \addlegendentry{Oracle (Things)}
    \addplot+[orange,solid,mark=none,ultra thick] table[x=IoU,y=Ours_things]{figures/figure2_final.txt};
    \addlegendentry{Ours (Things)}
    \addplot+[olive,solid,mark=none,ultra thick] table[x=IoU,y=MCN_things]{figures/figure2_final.txt};
    \addlegendentry{MCN~\cite{luo2020multi} (Things)}
    \addplot+[blue,dashed,mark=none,ultra thick] table[x=IoU,y=Oracle_stuff]{figures/figure2_final.txt};
    \addlegendentry{Oracle (Stuff)}
    \addplot+[blue,solid,mark=none,ultra thick] table[x=IoU,y=Ours_stuff]{figures/figure2_final.txt};
    \addlegendentry{Ours (Stuff)}
    \end{axis}
\end{tikzpicture}}
  \subcaption{Things and stuff categories}
  \label{fig:things_stuff}
\end{subfigure}
\begin{subfigure}[t]{.33\linewidth}
  \centering
      \resizebox{0.95\linewidth}{!}{%
  \begin{tikzpicture}[/pgfplots/width=1.45\linewidth, /pgfplots/height=1.45\linewidth]
    \begin{axis}[
                 ymin=0,ymax=1,xmin=0,xmax=1,
        		 xlabel=IoU,
        		 ylabel=Recall@IoU,
         		 xlabel shift={-2pt},
        		 ylabel shift={-3pt},
		         font=\small,
		         axis equal image=true,
		         enlargelimits=false,
		         clip=true,
        	     grid style=solid, grid=both,
                 major grid style={white!85!black},
        		 minor grid style={white!95!black},
		 		 xtick={0,0.1,...,1.1},
                 xticklabels={0,.1,.2,.3,.4,.5,.6,.7,.8,.9,1},
        		 ytick={0,0.1,...,1.1},
                 yticklabels={0,.1,.2,.3,.4,.5,.6,.7,.8,.9,1},
         		 minor xtick={0,0.02,...,1},
		         minor ytick={0,0.02,...,1},
        		 legend style={at={(0.05,0.05)},
                 		       anchor=south west},
                 legend cell align={left}]
    \addplot+[orange,dashed,mark=none,ultra thick] table[x=IoU,y=Oracle_singulars]{figures/figure3_final.txt};
    \addlegendentry{Oracle (Singulars)}
    \addplot+[orange,solid,mark=none,ultra thick] table[x=IoU,y=Ours_singulars]{figures/figure3_final.txt};
    \addlegendentry{Ours (Singulars)}
    \addplot+[blue,dashed,mark=none,ultra thick] table[x=IoU,y=Oracle_plurals]{figures/figure3_final.txt};
    \addlegendentry{Oracle (Plurals)}
    \addplot+[blue,solid,mark=none,ultra thick] table[x=IoU,y=Ours_plurals]{figures/figure3_final.txt};
    \addlegendentry{Ours (Plurals)}
    \end{axis}
\end{tikzpicture}}
  \subcaption{Singulars and plurals}
  \label{fig:singulars_plurals}
\end{subfigure}
   \caption{\textbf{Average Recall Curve} for our baseline method performance (a) compared to the oracle, and dissagregated into (b) things and stuff categories, and (c) singulars and plurals noun phrases.}
  \label{fig:average_recall_curve}
 \end{figure*}
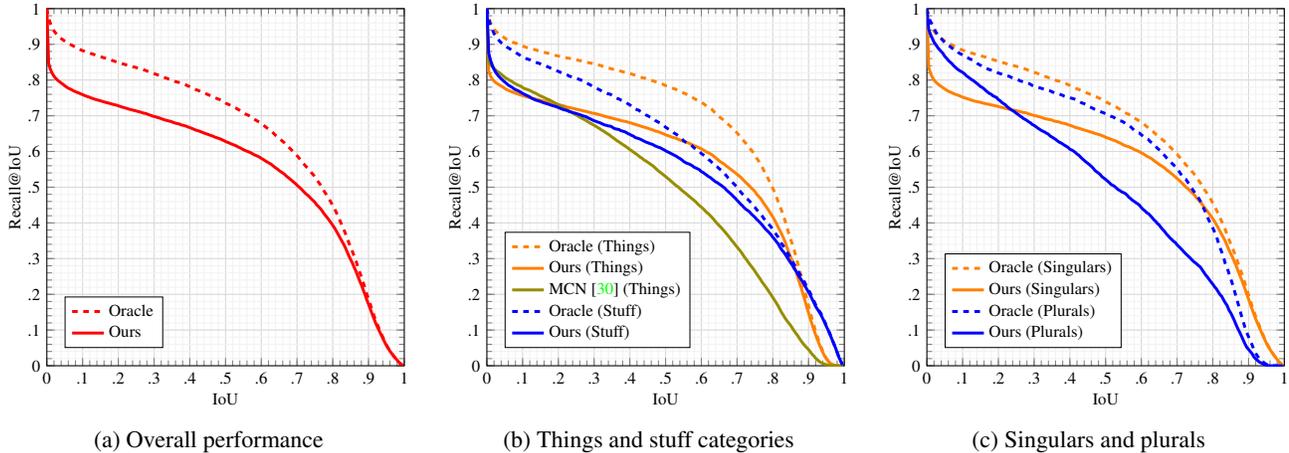

\begin{table*}[!htb]
\begin{subtable}{.5\linewidth}
  \centering
    \resizebox{0.87\linewidth}{!}{%
        \begin{tabular}{c@{\hspace{25mm}}ccc}
        \toprule
        \multirow{2}{*}{\textbf{Method}} & \multicolumn{3}{c}{\textbf{Average Recall}} \\ \cmidrule{2-4}
         & \textbf{things + stuff} & \textbf{things} & \textbf{stuff} \\ \midrule
        Oracle & 64.4 & 67.3 & 60.4  \\
        Ours & 55.4 & 56.2 & 54.3 \\ 
        MCN\cite{luo2020multi} & - & 48.2 & - \\\bottomrule
        \end{tabular}%
    }
    \vspace{1mm}
    \caption{Things and stuff categories.}
    \label{tab:things_stuff_results}
\end{subtable}%
\begin{subtable}{.5\linewidth}
  \centering
  \resizebox{0.9\linewidth}{!}{%
      \begin{tabular}{c@{\hspace{10mm}}ccc}
        \toprule
        \multirow{2}{*}{\textbf{Method}} & \multicolumn{3}{c}{\textbf{Average Recall}} \\ \cmidrule{2-4}
         & \textbf{singulars + plurals} & \textbf{singulars} & \textbf{plurals} \\ \midrule
        Oracle & 64.4 & 64.8 & 60.7 \\
        Ours & 55.4 & 56.2 & 48.8 \\ \bottomrule
        \end{tabular}%
  }
    \vspace{1mm}
    \caption{Singulars and plurals noun phrases.}
    \label{tab:singulars_plurals_results}
\end{subtable} 
\caption{\textbf{Results of our method} for \taskname{} task compared to the oracle performance, disaggregated into (a) things and stuff categories, and (b) singulars and plurals noun phrases.}
\label{tab:results}
\end{table*}

\begin{figure*}[!ht]
\begin{center}
   \includegraphics[width=0.89\linewidth]{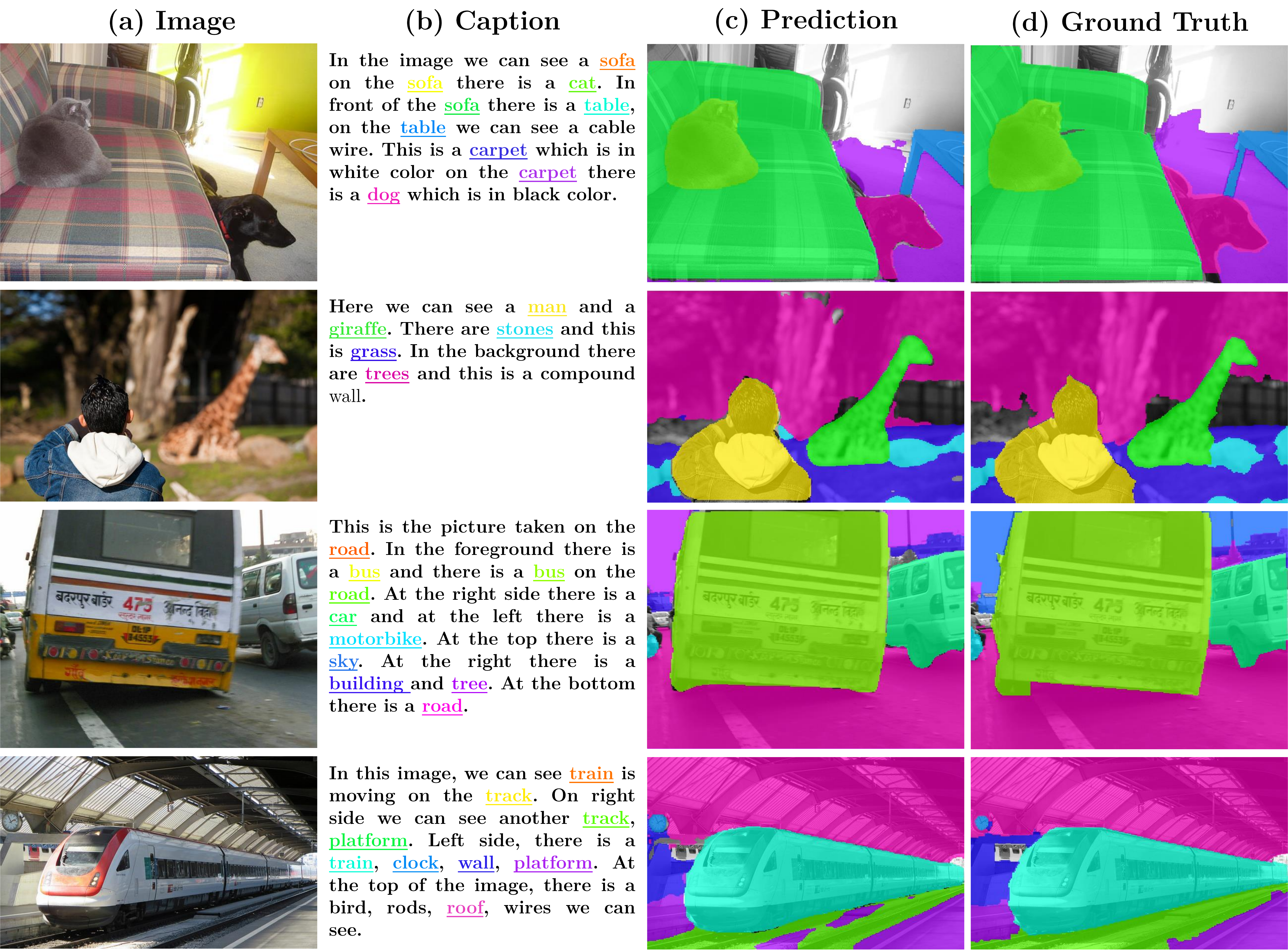}
\end{center}
\caption{\textbf{Qualitative results for \taskname{}.} Example predictions of our baseline in the validation split of our benchmark.
The inputs are the image (a) and the caption without highlighted noun phrases (b).
The outputs are a set of noun phrases in the caption (b), each with a corresponding region in the predicted segmentation (c).
    (d) shows the ground-truth panoptic segmentations.
More qualitative results can be found in the Supplementary Material.
}
    \label{fig:qualitative}
\end{figure*}

\mypar{Implementation details:} 
We use a Panoptic Feature Pyramid Network~\cite{kirillov2019panoptic} (FPN) with a ResNet-101~\cite{resnet} backbone pre-trained on \coco{}~\cococite{} with $3\times$ schedule using the official implementation~\cite{detectron2}.
Parameters in Panoptic FPN are fixed.
For stuff proposals, features correspond to the ones extracted from the semantic segmentation branch after combining the features extracted from each FPN level and before the final upsampling.
For things categories, they are the mask features in the instance segmentation branch.
The visual single modality transformer has \num{3} attention layers.
We use the pre-trained BERT ``base'' model~\cite{devlin2018bert} to generate the single-modality representation.
The cross-modality transformer has \num{5} attention layers.
During training, we performed a multi-label classification task to include several regions when the noun phrase is plural.
Additionally, for plurals in the final prediction we consider all the panoptic segmentation region proposal with a match score greater than \num{0.1}.
We train our method on an NVIDIA Quadro RTX 8000 GPU for \num{25} epochs with an initial learning rate of \num{0.00001}, \num{60} images per batch and use an Adam~\cite{adam} optimizer with standard parameters.

\section{Experiments}\label{sec:experiments}
\mypar{Experimental setup:}
We use our \taskname{} benchmark with the standard training and validation splits of the \coco{} dataset.
We only include the images for which our annotation transfer process assigns at least one noun phrase of the Localized Narrative caption to a panoptic segmentation region.
The final splits have \num{133103} localized narratives for train and \num{8533} for validation.

\mypar{Quantitative results:}
Table~\ref{tab:results} and Figure~\ref{fig:overall} show the quantitative results of our method using the proposed Average Recall evaluation metric (Sec.~\ref{subsec:metrics}).
Our method's relative performance is \num{86.0}\% with a drop of only \num{9.0} absolute Average Recall points with respect to the oracle performance given by the base panoptic segmentation method (\num{55.4} \vs{} \num{64.4}).
We calculate this oracle by choosing as prediction the panoptic segmentation proposal with the highest IoU with the ground-truth annotation.
Therefore, the oracle version of our method assumes a perfect match between segments and noun phrases, and only measures segmentation errors. 
These results suggest that our proposed baseline is very strong and a good starting point to promote research in \taskname{}.
However, our proposed baseline method is limited by the quality of the proposals' segmentations and the recall of the base panoptic segmentation method.

Furthermore, to compare our baseline for \taskname{} with respect to a visual grounding state-of-the-art method, Multi-task Collaborative Network~\cite{luo2020multi} (MCN).
We generalize our task and make a natural extension of RES task.
For this purpose, we split the captions into phrases and include as ground truth all segmentation regions that match a noun phrase within each sentence.
Since this method was developed in the scope of RES task, we just include things categories.
This state-of-the-art method for RES follows and bottom-up approach and leverages the complementary relationship between REC and RES tasks by using their properties to benefit the other task.
MCN achieves a performance of \num{48.2} Average Recall.
This result is comparable with the performance of our disaggregated method in only things, which shows that our baseline method better address the problem of natural language visual grounding.
We hypothesize that approaching this task from a fine-grained formulation allows the model to easily abstract information using the semantics of each word.

By disaggregating the performance of our baseline method into \emph{things} and \emph{stuff} (Tab.~\ref{tab:things_stuff_results}, Fig.~\ref{fig:things_stuff}),
it is possible to observe that,
even though our method's performance for \emph{stuff} categories is higher,
the drop with respect to the oracle performance for \emph{things} categories is \num{5.0} absolute Average Recall points greater.
We hypothesize that this is due to (i) greater variations in position and appearance in \emph{things} categories.
These results suggest that aligning noun phrases and panoptic region proposals gets more difficult as the intra-class visual variability increases.
Also, (ii) there can be multiple instances in an image of the same object category in the case of \emph{things}, as opposed to \emph{stuff} categories, so the model fails to disambiguate between the instances of that category and select the one to which the noun phrase refers in the caption's context.

We also disaggregate the performance on singular and plural noun phrases (Tab.~\ref{tab:singulars_plurals_results}, Fig.~\ref{fig:singulars_plurals}).
The drop in performance for plural noun phrases indicates that our model does not retrieve all the instances of the object to which language refers.
We attribute this limitation to the trace not exhaustively intersecting all the instances in many cases,
which spreads to our annotations and ends up affecting the results.

Figure~\ref{fig:qualitative} shows some qualitative results of our baseline method compared to the ground-truth panoptic segmentations of our benchmark.
More qualitative results can be found in the Supplementary Material.

\section{Conclusions}\label{sec:conclusions}
In this work, we present  \taskname{}, a new formulation for the natural language visual grounding problem that aims at producing a panoptic segmentation that grounds densely each noun phrase in the caption.
This version of the problem (i) maintains the finest granularity over language by visually grounding noun phrases,
while (ii) including a spatially detailed visual grounding with segmentations,
and (iii) incorporating all panoptic categories to leverage the intrinsic semantic information in visual scenes.
We establish a strong experimental framework for the study of this task,
including new annotations and evaluation metrics.
We design an algorithm to transfer Localized Narratives’ visual grounding to specific regions in the panoptic segmentations of the \coco{} dataset.
Furthermore, we propose a strong baseline method to serve as stepping stone for future work.
The formulation of \taskname{}, its benchmark and experimental framework will push the envelope further in
the development of fine and general natural language visual grounding methods.
In turn, advances in this task will impact solutions for other problems on the intersection of vision and language. 

{\small
\bibliographystyle{ieee_fullname}
\bibliography{grounding}

\begin{thebibliography}{10}\itemsep=-1pt

\bibitem{adelson2001seeing}
Edward~H Adelson.
\newblock On seeing stuff: the perception of materials by humans and machines.
\newblock In {\em Human vision and electronic imaging VI}, volume 4299, pages
  1--12. International Society for Optics and Photonics, 2001.

\bibitem{antol2015vqa}
Stanislaw Antol, Aishwarya Agrawal, Jiasen Lu, Margaret Mitchell, Dhruv Batra,
  C~Lawrence Zitnick, and Devi Parikh.
\newblock {VQA}: Visual question answering.
\newblock In {\em ICCV}, 2015.

\bibitem{nltk}
Steven Bird, Ewan Klein, and Edward Loper.
\newblock {\em Natural Language Processing with Python}.
\newblock O'Reilly Media, first edition, 2009.

\bibitem{chen2019see}
Ding-Jie Chen, Songhao Jia, Yi-Chen Lo, Hwann-Tzong Chen, and Tyng-Luh Liu.
\newblock See-through-text grouping for referring image segmentation.
\newblock In {\em ICCV}, 2019.

\bibitem{chen2015microsoft}
Xinlei Chen, Hao Fang, Tsung-Yi Lin, Ramakrishna Vedantam, Saurabh Gupta, Piotr
  Doll{\'a}r, and C~Lawrence Zitnick.
\newblock Microsoft {COCO} captions: Data collection and evaluation server.
\newblock {\em arXiv:1504.00325}, 2015.

\bibitem{chen2019referring}
Yi-Wen Chen, Yi-Hsuan Tsai, Tiantian Wang, Yen-Yu Lin, and Ming-Hsuan Yang.
\newblock Referring expression object segmentation with caption-aware
  consistency.
\newblock {\em arXiv:1910.04748}, 2019.

\bibitem{devlin2018bert}
Jacob Devlin, Ming-Wei Chang, Kenton Lee, and Kristina Toutanova.
\newblock {BERT}: Pre-training of deep bidirectional transformers for language
  understanding.
\newblock {\em arXiv:1810.04805}, 2018.

\bibitem{forsyth1996finding}
David~A Forsyth, Jitendra Malik, Margaret~M Fleck, Hayit Greenspan, Thomas
  Leung, Serge Belongie, Chad Carson, and Chris Bregler.
\newblock Finding pictures of objects in large collections of images.
\newblock In {\em International workshop on object representation in computer
  vision}, 1996.

\bibitem{resnet}
Kaiming He, Xiangyu Zhang, Shaoqing Ren, and Jian Sun.
\newblock Deep residual learning for image recognition.
\newblock In {\em CVPR}, pages 770--778, 2016.

\bibitem{heitz2008learning}
Geremy Heitz and Daphne Koller.
\newblock Learning spatial context: Using stuff to find things.
\newblock In {\em ECCV}, 2008.

\bibitem{hu2016segmentation}
Ronghang Hu, Marcus Rohrbach, and Trevor Darrell.
\newblock Segmentation from natural language expressions.
\newblock In {\em ECCV}, 2016.

\bibitem{hu2020bi}
Zhiwei Hu, Guang Feng, Jiayu Sun, Lihe Zhang, and Huchuan Lu.
\newblock Bi-directional relationship inferring network for referring image
  segmentation.
\newblock In {\em CVPR}, 2020.

\bibitem{huang2020referring}
Shaofei Huang, Tianrui Hui, Si Liu, Guanbin Li, Yunchao Wei, Jizhong Han, Luoqi
  Liu, and Bo Li.
\newblock Referring image segmentation via cross-modal progressive
  comprehension.
\newblock In {\em CVPR}, 2020.

\bibitem{hui2020linguistic}
Tianrui Hui, Si Liu, Shaofei Huang, Guanbin Li, Sansi Yu, Faxi Zhang, and
  Jizhong Han.
\newblock Linguistic structure guided context modeling for referring image
  segmentation.
\newblock In {\em ECCV}, 2020.

\bibitem{kazemzadeh2014referitgame}
Sahar Kazemzadeh, Vicente Ordonez, Mark Matten, and Tamara Berg.
\newblock Referitgame: Referring to objects in photographs of natural scenes.
\newblock In {\em EMNLP}, pages 787--798, 2014.

\bibitem{adam}
Diederik~P Kingma and Jimmy Ba.
\newblock Adam: A method for stochastic optimization.
\newblock {\em arXiv:1412.6980}, 2014.

\bibitem{kirillov2019panoptic}
Alexander Kirillov, Kaiming He, Ross Girshick, Carsten Rother, and Piotr
  Doll{\'a}r.
\newblock Panoptic segmentation.
\newblock In {\em CVPR}, 2019.

\bibitem{stanfordvp}
Jonathan Krause, Justin Johnson, Ranjay Krishna, and Li Fei{-}Fei.
\newblock A hierarchical approach for generating descriptive image paragraphs.
\newblock {\em arXiv:1611.06607}, 2016.

\bibitem{krishna2017visual}
Ranjay Krishna, Yuke Zhu, Oliver Groth, Justin Johnson, Kenji Hata, Joshua
  Kravitz, Stephanie Chen, Yannis Kalantidis, Li-Jia Li, David~A Shamma, et~al.
\newblock Visual genome: Connecting language and vision using crowdsourced
  dense image annotations.
\newblock {\em IJCV}, 123(1):32--73, 2017.

\bibitem{kuznetsova2020open}
Alina Kuznetsova, Hassan Rom, Neil Alldrin, Jasper Uijlings, Ivan Krasin, Jordi
  Pont-Tuset, Shahab Kamali, Stefan Popov, Matteo Malloci, Alexander
  Kolesnikov, et~al.
\newblock {The Open Images Dataset V4}.
\newblock {\em IJCV}, pages 1--26, 2020.

\bibitem{visualbert}
Liunian~Harold Li, Mark Yatskar, Da Yin, Cho{-}Jui Hsieh, and Kai{-}Wei Chang.
\newblock {VisualBERT}: {A} simple and performant baseline for vision and
  language.
\newblock {\em arXiv:1908.03557}, 2019.

\bibitem{li2020does}
Liunian~Harold Li, Mark Yatskar, Da Yin, Cho-Jui Hsieh, and Kai-Wei Chang.
\newblock What does bert with vision look at?
\newblock In {\em ACL}, 2020.

\bibitem{li2018referring}
Ruiyu Li, Kaican Li, Yi-Chun Kuo, Michelle Shu, Xiaojuan Qi, Xiaoyong Shen, and
  Jiaya Jia.
\newblock Referring image segmentation via recurrent refinement networks.
\newblock In {\em CVPR}, 2018.

\bibitem{li2017tracking}
Zhenyang Li, Ran Tao, Efstratios Gavves, Cees~GM Snoek, and Arnold~WM
  Smeulders.
\newblock Tracking by natural language specification.
\newblock In {\em CVPR}, 2017.

\bibitem{mscoco}
Tsung-Yi Lin, Michael Maire, Serge Belongie, James Hays, Pietro Perona, Deva
  Ramanan, Piotr Doll{\'a}r, and C~Lawrence Zitnick.
\newblock Microsoft {COCO}: Common objects in context.
\newblock In {\em ECCV}, 2014.

\bibitem{liu2017recurrent}
Chenxi Liu, Zhe Lin, Xiaohui Shen, Jimei Yang, Xin Lu, and Alan Yuille.
\newblock Recurrent multimodal interaction for referring image segmentation.
\newblock In {\em ICCV}, 2017.

\bibitem{liu2019learning}
Daqing Liu, Hanwang Zhang, Feng Wu, and Zheng-Jun Zha.
\newblock Learning to assemble neural module tree networks for visual
  grounding.
\newblock In {\em ICCV}, 2019.

\bibitem{vilbert}
Jiasen Lu, Dhruv Batra, Devi Parikh, and Stefan Lee.
\newblock {ViLBERT}: Pretraining task-agnostic visiolinguistic representations
  for vision-and-language tasks.
\newblock {\em arXiv1908.02265}, 2019.

\bibitem{luo2020cascade}
Gen Luo, Yiyi Zhou, Rongrong Ji, Xiaoshuai Sun, Jinsong Su, Chia-Wen Lin, and
  Qi Tian.
\newblock Cascade grouped attention network for referring expression
  segmentation.
\newblock In {\em ACM Multimedia}, 2020.

\bibitem{luo2020multi}
Gen Luo, Yiyi Zhou, Xiaoshuai Sun, Liujuan Cao, Chenglin Wu, Cheng Deng, and
  Rongrong Ji.
\newblock Multi-task collaborative network for joint referring expression
  comprehension and segmentation.
\newblock In {\em CVPR}, 2020.

\bibitem{refexp}
Junhua Mao, Jonathan Huang, Alexander Toshev, Oana Camburu, Alan~L. Yuille, and
  Kevin Murphy.
\newblock Generation and comprehension of unambiguous object descriptions.
\newblock {\em arXiv:1511.02283}, 2015.

\bibitem{mao2016generation}
Junhua Mao, Jonathan Huang, Alexander Toshev, Oana Camburu, Alan~L Yuille, and
  Kevin Murphy.
\newblock Generation and comprehension of unambiguous object descriptions.
\newblock In {\em CVPR}, 2016.

\bibitem{margffoy2018dynamic}
Edgar Margffoy-Tuay, Juan~C P{\'e}rez, Emilio Botero, and Pablo Arbel{\'a}ez.
\newblock Dynamic multimodal instance segmentation guided by natural language
  queries.
\newblock In {\em ECCV}, 2018.

\bibitem{miller1998wordnet}
George~A Miller.
\newblock {\em WordNet: An electronic lexical database}.
\newblock MIT press, 1998.

\bibitem{plummer17ijcv}
Bryan~A. Plummer, Liwei Wang, Christopher~M. Cervantes, Juan~C. Caicedo, Julia
  Hockenmaier, and Svetlana Lazebnik.
\newblock {Flickr30k Entities}: Collecting region-to-phrase correspondences for
  richer image-to-sentence models.
\newblock {\em IJCV}, 123(1):74--93, 2017.

\bibitem{pont2020connecting}
Jordi Pont-Tuset, Jasper Uijlings, Soravit Changpinyo, Radu Soricut, and
  Vittorio Ferrari.
\newblock Connecting vision and language with localized narratives.
\newblock In {\em ECCV}, 2020.

\bibitem{qiu2019referring}
Shuang Qiu, Yao Zhao, Jianbo Jiao, Yunchao Wei, and Shikui Wei.
\newblock Referring image segmentation by generative adversarial learning.
\newblock {\em IEEE Transactions on Multimedia}, 22(5):1333--1344, 2019.

\bibitem{rong2019unambiguous}
Xuejian Rong, Chucai Yi, and Yingli Tian.
\newblock Unambiguous scene text segmentation with referring expression
  comprehension.
\newblock 29:591--601, 2019.

\bibitem{selvaraju2019taking}
Ramprasaath~R Selvaraju, Stefan Lee, Yilin Shen, Hongxia Jin, Shalini Ghosh,
  Larry Heck, Dhruv Batra, and Devi Parikh.
\newblock Taking a hint: Leveraging explanations to make vision and language
  models more grounded.
\newblock In {\em ICCV}, 2019.

\bibitem{conceptual_captions}
Piyush Sharma, Nan Ding, Sebastian Goodman, and Radu Soricut.
\newblock Conceptual captions: A cleaned, hypernymed, image alt-text dataset
  for automatic image captioning.
\newblock In {\em ACL}, 2018.

\bibitem{shi2018key}
Hengcan Shi, Hongliang Li, Fanman Meng, and Qingbo Wu.
\newblock Key-word-aware network for referring expression image segmentation.
\newblock In {\em ECCV}, 2018.

\bibitem{vlbert}
Weijie Su, Xizhou Zhu, Yue Cao, Bin Li, Lewei Lu, Furu Wei, and Jifeng Dai.
\newblock {VL-BERT}: Pre-training of generic visual-linguistic representations.
\newblock {\em arXiv:1908.08530}, 2020.

\bibitem{suhr2017corpus}
Alane Suhr, Mike Lewis, James Yeh, and Yoav Artzi.
\newblock A corpus of natural language for visual reasoning.
\newblock In {\em ACL}, 2017.

\bibitem{suhr2018corpus}
Alane Suhr, Stephanie Zhou, Ally Zhang, Iris Zhang, Huajun Bai, and Yoav Artzi.
\newblock A corpus for reasoning about natural language grounded in
  photographs.
\newblock {\em arXiv:1811.00491}, 2018.

\bibitem{vaswani2017attention}
Ashish Vaswani, Noam Shazeer, Niki Parmar, Jakob Uszkoreit, Llion Jones,
  Aidan~N. Gomez, Lukasz Kaiser, and Illia Polosukhin.
\newblock Attention is all you need.
\newblock In {\em NeurIPS}, 2017.

\bibitem{detectron2}
Yuxin Wu, Alexander Kirillov, Francisco Massa, Wan-Yen Lo, and Ross Girshick.
\newblock Detectron2.
\newblock \url{https://github.com/facebookresearch/detectron2}, 2019.

\bibitem{ye2020dual}
Linwei Ye, Zhi Liu, and Yang Wang.
\newblock Dual convolutional lstm network for referring image segmentation.
\newblock {\em IEEE Transactions on Multimedia}, 22(12):3224--3235, 2020.

\bibitem{ye2019cross}
Linwei Ye, Mrigank Rochan, Zhi Liu, and Yang Wang.
\newblock Cross-modal self-attention network for referring image segmentation.
\newblock In {\em CVPR}, 2019.

\bibitem{young14tacl}
Peter Young, Alice Lai, Micah Hodosh, and Julia Hockenmaier.
\newblock From image descriptions to visual denotations: New similarity metrics
  for semantic inference over event descriptions.
\newblock {\em TACL}, 2:67--78, 2014.

\bibitem{young2014image}
Peter Young, Alice Lai, Micah Hodosh, and Julia Hockenmaier.
\newblock From image descriptions to visual denotations: New similarity metrics
  for semantic inference over event descriptions.
\newblock {\em TACL}, 2:67--78, 2014.

\bibitem{yu2018mattnet}
Licheng Yu, Zhe Lin, Xiaohui Shen, Jimei Yang, Xin Lu, Mohit Bansal, and
  Tamara~L Berg.
\newblock Mattnet: Modular attention network for referring expression
  comprehension.
\newblock In {\em CVPR}, 2018.

\bibitem{zheng2020cross}
Chen Zheng, Quan Guo, and Parisa Kordjamshidi.
\newblock Cross-modality relevance for reasoning on language and vision.
\newblock {\em arXiv:2005.06035}, 2020.

\bibitem{zhou2017scene}
Bolei Zhou, Hang Zhao, Xavier Puig, Sanja Fidler, Adela Barriuso, and Antonio
  Torralba.
\newblock Scene parsing through ade20k dataset.
\newblock In {\em CVPR}, 2017.

\end{thebibliography}
}

\end{document}